# Minimum Message Length Clustering Using Gibbs Sampling

Ian Davidson [1]
inpd@hotmail.com
Monash University, Clayton Road, Victoria Australia

## Abstract

The K-Means and EM algorithms are popular in clustering and mixture modeling due to their simplicity and ease of implementation. However, they have several significant limitations. Both converge to a local optimum of their respective objective functions (ignoring the uncertainty in the model space), require the apriori specification of the number of classes/clusters, and are inconsistent. In this work we overcome these limitations by using the Minimum Message Length (MML) principle and a variation to the K-Means/EM observation assignment and parameter calculation scheme. We maintain the simplicity of these approaches while constructing a Bayesian mixture modeling tool that samples/searches the model space using a Markov Chain Monte Carlo (MCMC) sampler known as a Gibbs sampler. Gibbs sampling allows us to visit each model according to its posterior probability. Therefore, if the model space is multi-modal we will visit all modes and not get stuck in local optima. We call our approach multiple chains at equilibrium (MCE) MML sampling.

## 1 INTRODUCTION AND OVERVIEW

We will begin this paper by introducing the EM algorithm and K-means algorithm and illustrate their mechanisms to assign observations to classes and calculate class parameters. We then describe some of their inherent limitations. We introduce the MML principle and MCMC albeit briefly. We illustrate how by changing the basic EM and K-Means algorithms and using a MML estimator, we construct a sampler that explores the model space visiting models with a chance equal to their posterior probability. Our approach overcomes the previously mentioned limitations of K-Means and the EM algorithm. Furthermore, these algorithms effectively ignore the uncertainty in the model space. Accepting and considering this uncertainty by visiting models according to their posterior probability has many benefits. One can sample from progressively less severe power transformations of the posterior an approach commonly known as simulated annealing that has been shown to find good local optima (Aarts and Korst 1989). By sampling directly from the posterior and collecting alternative explanations of the data, a better understanding of the model space with respect to the data is achieved. Similarly, if we wish to make predictions of some kind, we can sample from the posterior and perform Bayesian model averaging. Empirical results comparing our sampler against EM are presented next and we conclude by summarizing the contributions of this paper and describing related work.

## 2 THE EM AND K-MEANS CLUSTERING ALGORITHMS

The Expectation Maximization (EM) algorithm (Dempster et al 1977) and the K-Means clustering algorithm (MacQueen 1967) are two techniques for searching the model space. Both attempt to find the single best point estimator within the model space though it is well known that the definition of "best" varies between the two.

The EM algorithm in the classical inference setting attempts to find the maximum likelihood estimate (MLE) or in a Bayesian setting the maximum a posteriori (MAP) estimate. The K-Means algorithm aims to find the minimum distortion within each cluster for all clusters. Both algorithms consist of two fundamental steps:

1) *Observation assignment* step where the observations are assigned to classes based on class descriptions.

2) *Parameter re-estimation* step where the class descriptions are recalculated from observations assigned to them.

The two steps are repeated until convergence to a point estimator is achieved.

In the first step of the K-Means algorithm the observations are assigned *exclusively* to the "closest" class as defined by some distance metric. Euclidean distance is often used. In the EM algorithm an observation is assigned *partially* to each cluster, the portion of the observation assigned depending on how probable (or

---

likely) the class generated the observation.

In the second step both algorithms use the attribute values of the observations assigned to a class to recalculate class parameters. As K-Means uses exclusive assignment we recompute the estimates from only those observations that are assigned to the class. However, in the EM algorithm the class parameter estimates are calculated from each observation weighted by the proportion of the observation assigned to the class.

Let the $k$ classes partition the observations into the subsets $C_{1...k}$, the cluster centroids be represented by $w_{1...k}$ and the $n$ elements to cluster be $S_{1...n}$. The minimum distortion or vector quantization error (with constants removed) that the K-Means algorithm attempts to minimize is shown in equation ( 1 ). The mathematical trivial solution which minimizes this expression is to have a cluster for each observation.

$$Distortion = \sum_{j=1}^{k} \sum_{i=1}^{N} \sum_{S_i \in C_j} D(S_i, w_{Class(S_i)})^2 \quad (1)$$

where $D$ is some distance metric

The EM algorithm attempts to minimize the log loss which is precisely the local maximum of the likelihood that the model (the collection of classes) produced the data. The likelihood is shown in equation ( 2 ). The class description ($w_j$) is now a vector of probability distributions for each attribute for the $j^{th}$ class and $p_j$ is the probability of the $j^{th}$ class.

$$P(D \mid H) = \sum_{j=1}^{k} p_j \prod_{i=1}^{N} P(S_i \mid w_j) \quad (2)$$

Both algorithms converge to a local optimum of the respective functions they attempt to optimize. We shall refer to these functions as objective functions.

## 3 LIMITATIONS OF THE K-MEANS AND EM ALGORITHMS

We illustrate three limitations of the K-Means and EM algorithms:

1. The inconsistency of the estimators.
2. The estimators find the local optimum of their respective objective functions.
3. The estimators require the a-priori specification of the number of classes.

### 3.1 THE ESTIMATORS ARE INCONSISTENT

Consider a model space $\Theta_k$, which contains models of only $k$ classes including, $\theta_T$, the true model that generated the observations. Initially, there maybe only a small number of observations in our sample so $\theta_T$ is not the most probable model in the model space. If an estimator is consistent then we find that:

$$\lim_{n \to \infty} P(\theta_T) = 1, \text{ where } n \text{ is the number of observations} \quad (3)$$

That is, as the amount of data increases the probability of the true model approaches certainty. An inconsistent estimator does not have this property and instead we find that the true model is overlooked in favor of increasingly complex (more classes) models. Consider the objective functions of K-means and the EM algorithm in equations ( 1 ) and ( 2 ) respectively, the (trivial) optimal solution is to have a cluster for each observation. It is precisely these "biases" which leads the estimator to consistently favor increasingly (as more data is available) complicated models.

### 3.2 THE ALGORITHMS FINDS LOCAL OPTIMA OF THEIR OBJECTIVE FUNCTIONS

The vector quantization error and likelihoods are locally optimized by the K-means and EM algorithms respectively. Both algorithms perform a gradient ascent/descent of their objective functions and can therefore become stuck in local optima. Hence, the algorithms provide a local optimum of their objective function. For most interesting practical problems the error surface will contain many local optima (Gilks et al 1996).

### 3.3 THE ESTIMATORS REQUIRE A-PRIORI SPECIFICATION OF THE NUMBER OF CLASSES

Both the K-Means and EM algorithms require the apriori specification of the number of classes. The model space explored is all possible models with $k$ classes and effectively removes $k$ (an important unknown in intrinsic classification) from the problem. One can select a desirable range of $k$ and use the algorithm for each value within the range, but due to the algorithm finding only local optimum, this process would need to be completed many times to get the "best" model for each $k$. However, we cannot easily compare models obtained for different values of $k$. The distortion or likelihood for models with a large $k$ will have a greater potential to be better than those models with a small $k$. One can use various metrics such as Akaike Information Criterion or Bayesian Information Criterion to determine the number of classes. However, they have been shown to be less accurate than MML estimators for choosing the correct model space (Oliver et al 1996).

We are somewhat stuck. The algorithms require a specification of $k$ but we cannot compare the objective function across different values of $k$. We now introduce the MML principle and MCMC sampling which can in theory overcome these limitations.

## 4 THE MML PRINCIPLE

The process of inductive learning essentially abstracts, generalizes or *compresses* the observations into a model. Solomonoff first formally described the relationship between induction and compression in his seminal paper on inductive inference (Solomonoff 1964). He noted that a computer program (the theory) was a compressed version of its output (the observations). The best program/model is the shortest in length as it explains all current observations and being the smallest is the most general. The length of this program is the Kolmogorov complexity of the observations represented as a string. Unfortunately as is noted by Solomonoff (Solomonoff 1996) and Chaitin (Chaitin 1970) measures such as the Kolmogorov complexity are incomputable and hence applications of inference based on them impractical.

Wallace and Boulton independently of Solomonoff, Chaitin and Kolmogorov formulated and applied their MML principle in a series of papers on intrinsic classification (Wallace and Boulton 1968) (Boulton and Wallace 1970) (Boulton and Wallace 1973). Their principle uses Shannon information theory as a mechanism for compressing observations and hence overcomes the incomputability problem associated with Kolmogorov complexity.

MML inference involves constructing a two-part string which represents the observations that can be transmitted between a sender and receiver. The first part (which is received first) is the model or theory of the observations whilst the second part is the observations encoded with respect to the model. The best model has the shortest total (sum of both parts) message length. This involves an implicit trade off between model complexity and the model "fitting" the observations.

The principle can be re-stated in a Bayesian form (Wallace and Boulton 1975) (Wallace and Freeman 1987) with the length of the first part of the string used to calculate the Bayesian prior and the length of the second part the likelihood. The Bayesian posterior which is the primary concern of Bayesian inference is shown in equation ( 4 ).

$$P(\theta_i \mid D) = \frac{P(\theta_i).P(D \mid \theta_i)}{P(D)} = \frac{P(\theta_i \cap D)}{P(D)} \quad (4)$$

where: $P(\theta_i)$ is the prior probability of model $i$.
$P(D \mid \theta_i)$ is the likelihood.
$P(D)$ is the probability of the data.

Taking the negative logarithm of this expression yields:
$$-\ln P(\theta_i D) = -\ln P(\theta_i) - \ln P(D \mid \theta_i) + \ln(P(D)) \quad (5)$$

Information theory tells us that for a collection of mutually exclusive and exhaustive events, representing the events with code words (a unique concatenation of symbols) of length $-\log(P(event))$ results in a minimum length message. Therefore, by minimizing equation ( 5 ) we inherently maximize the posterior probability and identify the most probable model. Minimizing this equation involves searching for the model that gives the shortest message. The posterior probability can be approximated from the length of the two-part message since from equations ( 4 ) and ( 5 ):

$$P(\theta_i \cap D) = P(\theta_i).P(D \mid \theta_i)$$
$$= e^{-\ln(P(\theta_i)) - \ln(P(D \mid \theta_i))}$$
$$= e^{-(MessageLength_{Part1} + MessageLength_{Part2})} \quad (6)$$

and

$$P(\theta_i \mid D) \propto e^{-(MessageLength_{Part1} + MessageLength_{Part2})} \quad (7)$$
where $e$ is Euler's number, ln is the natural logarithm

For induction using discrete multi-state variables MML induction reduces to Bayes theorem. However, with continuous variables MML effectively discretises the parameter space into optimal regions thereby allowing a probability estimate (not a density) to be attached to each region. The MML estimate for a given induction problem is the representative model for the most probable region.

Consider encoding 500 observations from a Gaussian distribution with a mean of 0 and standard deviation of 1. For a particular sample the sample mean is approximately 0.044, the sample standard deviation 1.016, the $Range_\mu$ = [-5,5] and the $Range_\sigma$ = [0,5]. The highly probable regions that we obtain by using the MML principal are shown in Figure 1. The most likely adjacent region apart from the MML estimator containing region is 41 times less likely than it whilst the least probable region is some 2.8 million times less likely. As the number of attributes (dimensions) in the problem increases the posterior odds ratios of the adjacent regions will increase making them even less likely.

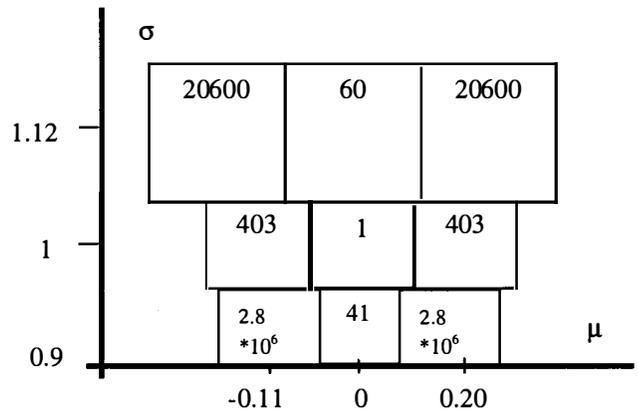

Figure 1: The posterior odds ratio of the MML regions containing and adjacent to the MML estimate for a 500 observation sample from the population $\mu$=0, $\sigma$=1

The message length calculations for mixture modeling that we use can be found elsewhere (Wallace and Boulton 1968) (Davidson 1998). We propose using the relationship between the message length and the posterior to construct a Markov chain whose stationary distribution is the posterior distribution. A stationary distribution is a probability distribution over all states (which are models in our case) which once reached persists forever.

## 5 MARKOV CHAIN MONTE CARLO SAMPLING

Markov chain Monte Carlo (MCMC) sampling was first postulated in the seminal paper of Metropolis et al (Metropolis et al 1953) as a method to model systems at thermodynamic equilibrium. Systems at thermodynamic equilibrium are computationally interesting for two primary reasons:

1) The probability of a system at equilibrium being in a specific state at a constant temperature is given by the well-known Boltzmann distribution (equation (8)).

2) If the temperature is reduced sufficiently slowly and equilibrium found at each temperature, the system will converge to its lowest energy state.

Both situations can be modeled as a Markov chain. The first situation can be expressed as:

$$P(X=i) = \frac{1}{N_O(c)} \cdot e^{\left(\frac{-f(i)}{kc}\right)}$$

$$where, N_O(c) = \sum_{j \in \Theta} e^{\left(\frac{-f(j)}{kc}\right)} \quad (8)$$

This equation states that the probability of the system being in a state $i$ (from a set $\Theta$) is dependent only on the energy of the specific state, $f(i)$ and the temperature $c$. We can relate the situation described by equation (8) to MML inference by allowing state $i$ to be a specific model and substituting the sum of the length of the two part message for $f(i)$, then equation (8) becomes:

$$P(X=i) = \frac{e^{-\left(\frac{MessLen(i)}{c}\right)}}{\sum_{j \in \Theta} e^{-\left(\frac{MessLen(j)}{c}\right)}} = \frac{e^{-\left(\frac{-\ln(P(H_i))+-\ln(P(D|H_i))}{c}\right)}}{\sum_{j \in \Theta} e^{-\left(\frac{-\ln(P(H_j))+-\ln(P(D|H_j))}{c}\right)}}$$

$$= \left(\frac{P(H_i).P(D|H_i)}{\sum_{j \in \Theta} P(H_j).P(D|H_j)}\right)^{1/c} = \left(\frac{P(H_i).P(D|H_i)}{P(D)}\right)^{1/c} \quad (9)$$

Therefore, by simulating a system at equilibrium (by constructing an appropriate Markov chain) and using the message length as an energy function the probability of being in a state (a particular model) is exactly its posterior probability at temperature 1.

The second reason systems at thermodynamic equilibrium are interesting was first computationally used by Kirkpatrick et al (Kirkpatrick et al 1983) in their method known as simulated annealing. They showed that by extending the thermodynamic analogy, combinatorially large NP hard problems could be handled. We will now introduce the popular Gibbs sampling approach for performing MCMC sampling.

## 6 GIBBS SAMPLING

If it is possible to generate conditional probability estimates of the posterior distribution, Gibbs sampling can be used easily. The algorithm has become popular since its use by Geman and Geman (Geman and Geman 1984). Consider the situation of a random variable, $X^{(t)}$, which represents the entire model. We can perform a conditional *asynchronous* update of each individual component within $X^{(t)}$ to derive the new component value. This effectively constructs a Markov chain whose stationary distribution is the posterior. The components are the individual problem unknowns we are trying to find estimates for, which in intrinsic classification are the class sizes, class descriptions and which observations are assigned to each class. In theory no particular order of component updating is required though usually each component is updated in sequence, a process known as a sweep. Updating a component assigns it a new value according to the probability distribution of the values for the component, conditional on all other component values. Formally the process for updating $n$ components at time $t$, in order can be described as:

Pick $X_1^{(t)}$ from $P(X_1| x_2^{(t-1)}, x_3^{(t-1)}, \ldots x_n^{(t-1)})$

Pick $X_i^{(t)}$ from $P(X_i| x_1^{(t)} \ldots x_{i-1}^{(t)}, x_{i+1}^{(t-1)} \ldots x_n^{(t-1)})$ (10)

Pick $X_n^{(t)}$ from $P(X_n| x_1^{(t)}, x_2^{(t)}, \ldots x_{n-1}^{(t)})$

If we wish to sample from the posterior we use the sampling process defined in equation (10). To perform annealing we simply raise these conditional distributions to the power of $c^{-1}$.

## 7 HOW A MCMC MML SAMPLER OVERCOMES THE LIMITATIONS OF K-MEANS AND THE EM ALGORITHM

The MML estimator we use can be viewed as attempting to minimize the Kullback-Leibler distance between the parameter estimates and the marginalized prior distribution of the data. Unlike the objective functions of the EM and K-Means algorithms the number of observations is not a factor, thus increasing the number of observations does not directly effect the objective function. The consistency of MML estimators has formally be shown (Barron and Cover 1991).

The K-Means and EM algorithms effectively perform gradient descent and therefore can become stuck in local optima. Colloquially, the probabilistic nature of the MCMC sampler ensures that the sampler will not get stuck in local optima, more formally it can be shown that a Gibbs sampler will converge to visiting the models with a chance equal to their posterior distribution.

As we have a probability estimate for each model we can compare models of different complexity. However, we must still sample according to the posterior both within model spaces of a fixed dimension and across model spaces of different dimensions.

Most MCMC sampling assumes a fixed dimension model space since:

1) Bayesian formulation of problems with continuous

parameters often provide posterior probability density estimates rather than probabilities for each model.

2) Metropolis's original work involved simulating particles in a fixed two-dimensional space.

Sampling from a model space of varying dimension can be overcome by at least two methods. Firstly by effectively increasing the number of dimensions for each model so that it is always a constant and secondly by converting the probability densities to probability estimates. There have been several techniques (jump diffusion and reverse jump dynamics) recently developed that allow sampling across a model space of varying dimension. Both have their limitations and are quite involved. Reverse jump dynamics uses a combination of stochastic and deterministic updating mechanisms to effectively increase the dimension of the entire model space to a constant. Whilst jump diffusion still uses probability densities, the mechanism for sampling effectively calculates a probability estimate for each value of $k$ which can be used to move between dimensions.

Our sampler also overcomes other limitations specific to either K-Means or the EM algorithms which we have not discussed. For example K-Means is not invariant to non-linear re-parameterization of the data while both EM and our clustering tool are.

We shall now describe a Gibbs sampler that overcomes the previously mentioned limitations by using the MML principle and an algorithm quite similar to the K-Means and EM algorithms.

## 8 CREATING A GIBBS SAMPLER WITH A MML CLUSTERING TOOL

A Markov chain sampling from its stationary distribution is at equilibrium. If we have *multiple* chains at equilibrium exploring different sub-spaces then jumping between them in an appropriate manner would overcome the problem of varying state space dimension. As we are only simulating the chains in discrete time, we can stop the chain and re-start it at a later time and still be sampling from the stationary distribution. We have an ensemble of $n$ chains at equilibrium each sampling from a part of the model space with a different dimension. At any given instance, time is being advanced in only one of the chains which we call the *active chain*. The remaining chains are not advanced until they are chosen to become the active chain.

We will first illustrate our approach to create one of these chains that samples from a model space of a fixed number of components and then discuss jumping between multiple chains. We use uniform priors for the number of classes, class sizes and parameter estimates. Our message length calculations for mixture modeling are similar to those of others (Boulton and Wallace 1973) and we note that other message length calculations could have been used instead.

### 8.1 SIMULATING A MARKOV CHAIN FOR A FIXED NUMBER OF COMPONENTS

Our approach to simulating a chain for a fixed number of components is very similar to the EM and K-Means algorithms. In the first step, we assign observations exclusively to one class by a random experiment according to the observation's normalized posterior probabilities. The posterior probabilities for an observation are obtained by determining the change in message length by assigning an observation to difference classes and from using equation ( 7 ). In a two-class problem, suppose message lengths of 200 nits resulted if an observation were to be assigned to class 1 and 200.5 nits if assigned to class 2. Then the normalized posterior probability of the observation belonging to the classes would be .62 and 0.38 respectively. By tossing a biased coin we would assign the observation exclusively to one of the classes.

In the second step, we calculate the parameter estimates based on these exclusive assignments and the process repeats as is in K-Means and the EM algorithms. However our sampler does not converge to a point estimate. The pseudo code for this approach is shown in Figure 2.

```
For every observation
    For every class
        Calculate and store the
        messagelength if this observation
        were assigned to this class
    EndFor
    From equation ( 7 ) calculate the
    normalized posterior probabilities
    Randomly assign the observation to a
    class according to its posterior
End For

Re-calculate the parameter estimates of
the classes from the exclusive
assignments
```

Figure 2: One Sweep of Gibbs sampling in a fixed dimensional space.

The process we have just described is a form of Gibbs sampling. There are three sets of unknowns in our formulation of intrinsic classification: the observations' class assignments, $S_j$, $j=1...N$, the classes' parameter estimates (assume Gaussian distributed attributes), $\{\mu_{k,m}, \sigma_{k,m}\}$, $k=1... T$, $m=1... M$ and the class relative abundances, $p_k$, $k = 1...T$. Recall, that to perform Gibbs sampling we need to update each of the unknowns conditional on all other unknowns.

Randomly assigning observations to classes updates $P(S_j \mid p, \mu, \sigma)$, for $j= 1...N$, according to the posterior. However, we must also update the remaining two sets of unknowns, $P(p \mid \mu, \sigma, S_j: j = 1 ... N )$ and $P(\mu_{k,m}, \sigma_{k,m} \mid p_k, S_j: j = 1 ... N)$, for $k = 1 ... T$. The first expression simplifies to $P(p \mid$ $S_j$: $j = 1 \ldots N$ ) due to independence between the class sizes and parameter estimates. Similarly, the second expression simplifies to $P(\mu_{k,m}, \sigma_{k,m} \mid S_j: j = 1 \ldots N)$. However, we explicitly *calculate* the values of class abundance and class parameter estimates from $S_j$. How does this approach update these unknowns according to their posterior distribution ?

Though we haven't randomly sampled from their posterior distributions per se, this process is approximated. The size and shape of the MML region that the MML estimate belongs to, is similar to the region of expected error of estimating the parameters of the probability distribution from the sample (Wallace and Boulton 1968,Appendix). Put simply, the MML estimate we obtain by calculating the class parameter estimates from the observations is the only class parameters with a large probability. All other estimates have a probability near zero as is illustrate in the univariate situation in Figure 1.

Let us further discuss this by considering Figure 1 which is the equivalent situation to class parameter estimation if we consider the sample to be the observations within the class. There are infinitely many parameter estimates for the class. However, MML reduces this to a finite number by optimally discritising the parameter space into regions. Furthermore, of the regions, only the one containing the MML estimate effectively has a non-zero probability. Therefore the MML estimates of the class we *calculated* are the only estimates that would have been chosen if we randomly updated the class parameter estimates according to their posterior distribution. A similar situation exists when updating the class weights/sizes.

The approach we have just described removes the first two limitations of K-Means and EM. To remove the third limitation (the requirement to specify the number of components) we must expand the model space to contain models of varying dimensions. However, this requires moving between model spaces of different complexity and to achieve this we need to modify our base approach.

## 8.2   JUMPING BETWEEN MODEL SUB-SPACES

As we have many chains at equilibrium but only advancing (or sampling) from one at any given time we must jump between them in an appropriate manner to ensure that we are sampling according to the posterior. The posterior probability for the entire state space our sampler explores (limited to at most $K$ classes) is shown in equation ( 11 ).

$$P(\Theta \mid D) = \sum_{k=1}^{K} P(\Theta_k \mid D) \propto \sum_{k=1}^{K} \sum_{j} P(\theta_j).P(D \mid \theta_j) \quad (11)$$
$$\theta_j \in \Theta_k$$

To maintain sampling according to the posterior we must visit the model sub-space of $k$ components with probability $P(\Theta_k \mid D)$. We can explicitly calculate or approximate $P(\Theta_k \mid D)$ but to do so accurately is difficult. However, we only need relative estimates for different values of $k$ which makes the task easier. Once a chain has reached equilibrium we can use the message lengths of the observations from the chain to calculate an approximation of $P(\Theta_k \mid D)$ by using a population estimation approach to determine the number of highly probable models in the sub-space.

We are not concerned with relatively improbable models as given sufficient data they will not contribute much to the probability of the entire sub-space. The population estimation technique we use follows. We simulate each chain for $M$ iterations after it has reached equilibrium. For each chain we create adjacent, non-overlapping bins $(b_{1\ldots j})$ that are one nit in width that span from the shortest through to the longest message length (rounded up to the nearest nit) of the $M$ observations. We place each of the $M$ observations into one bin based on its message length so that we have the number of visits $(v_{1\ldots j})$ to each bin.

It is determining a good estimate to the true number of unique models in each bin $(m_{1\ldots j})$ that will allow us to calculate an approximation to $P(\Theta_k \mid D)$. The approximation is shown in equation ( 12 ).

$$P(\Theta_k \mid D) \approx \sum_{j=1}^{J} m_j . e^{-ML(b_j)} \quad (12)$$

where $ML(b_j)$ is the message length at the centre of bin $b_j$

Of course each visit does not correspond to a unique model in the bin as the same model may have been visited many times. The value, $m'_j$ is the estimate of the number of unique models that according to our sampler are in the $j^{th}$ bin. If $m_j$ is the actual number of unique models within the boundaries of $b_j$ then the chance that our sampler did not visit a particular unique model is:

$$\left(1 - \frac{1}{m_j}\right)^{v_j} \text{, if } m'_j \approx m_j \text{ then,}$$

$$m_j \left(1 - \left(1 - \frac{1}{m_j}\right)^{v_j}\right) \approx m'_j,$$

substitute $m'_j$ with $m_j$

$$\therefore m_j \left(1 - \left(1 - \frac{1}{m_j}\right)^{v_j}\right) - m_j \approx 0 \quad (13)$$

By solving equation ( 13 ) for $m_j$ for each bin, for progressively increasing message lengths (decreasing probability) until the values of $m_j$ and $m'_j$ differ by more than 1 we can use our estimates of $m_j$ in equation ( 12 ).

As we are only interested in relative probabilities of each sub-space we can sample a fixed number of times ($M$) from each sub-space to get approximation of $P(\Theta_k \mid D)$.

After the estimations for each sub-space are completed then we can normalize them and by drawing a random number between 0 and 1 jump to the next active chain (sub-space). The pseudo code to generate the posterior estimates for each sub-space is shown in Figure 3

```
// To begin, each chain is at equilibrium
Generate M observations from each of the j
chains at equilibrium.
```

```
For each of the j chains and their M
observations

// Create bins of one nit in width
    Bin_Max = ceil(Longest Message Length)
    Bin_Min = floor(Shortest Message Length)
    Create (Bin_Max-Bin_Min) counters (V_j)

    For j = 1 to (Bin_Max - Bin_Min) step 1
        V_j = 0
        MessageLength_j = Bin_Min + j - 0.5

// Count how many of the M observations
// belong in each bin counter.

        For each of the M observations
            If Observation's message length
            is within MessageLength_j±0.5 then
                V_j++
        EndFor
    EndFor

// Calculate posterior for jth chain
Posterior_j = 0

// Cycle through each bin
    For j = 1 to (Bin_Max - Bin_Min) step 1
        Solve equation ( 13 )
        If no solution then exit loop
        Posterior_j += m_j.e^{-MessageLength_j}
    End For
Endfor
```

Figure 3: Pseudo code for obtaining posterior estimates for each sub-space.

As more observations are obtained from each sub-space we can use a similar process to update our estimates of the posterior probability of that sub-space.

## 9 EMPIRICAL RESULTS

A sampler visiting each model according to its posterior probability, will, when used with an annealing type heuristic hopefully converge to a good local optima. As we allow movement to less probable models (effectively moving up a hill) we expect to outperform gradient ascent/descent algorithms like EM and K-Means. We compare our MCE sampler against a search algorithm based on the EM algorithm (in a maximum likelihood context) that systematically changes the number of classes (value of $k$). We refer to this search as EM search.

We tried the EM search on a six Gaussian variate problem consisting of six components all with means of 0 and standard deviations of 0.5 except for component $i$ whose $i^{th}$ attribute has a mean of 1. That is, $\mu_{1...6,1..6}=0$ except $\mu_{i,i}=1$, $\sigma_{1..6,1..6}=0.5$. We generate 500 data points from each component for a total of 3000 observations. This is a difficult problem as there is considerable overlap between the classes. Figure 4 illustrates the same problem but for only two dimensional space and two components.

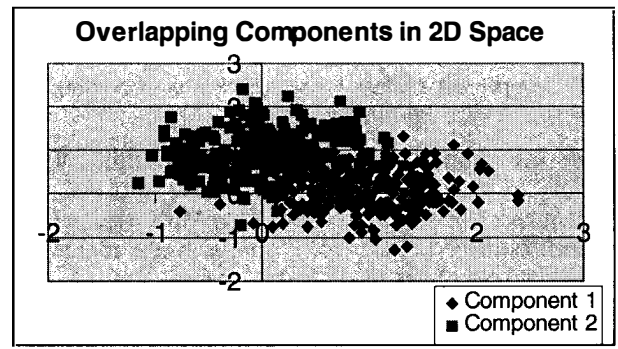

Figure 4: 500 Observations from component 1 ( N(1,0.5), N(0,0.5) ) and component 2 ( N(0,0.5), N(1,0.5) )

We continually ran randomly restarted EM search for an one-hour period and told it to start initially with 5, 10, 15 and 20 classes. The unencoded data is approximately 58400 nits. The best model found by EM search resulted in a message length of 58031.3 nits

We found that EM search finds the correct number of classes but the class populations and parameter estimates are not accurate. This is because there is quite a deal of overlap amongst the classes and there are many local optima in the posterior which the greedy EM search heuristic gets attracted to. Another factor which resulted in EM search not fairing well on this problem is because it was not initially told the correct number of classes. However, during an overnight run EM search finds a model of length 58013 nits whose parameter estimates are closer to the generation mechanism's. If we specify EM search to only look in the 6-class sub-space then within one hour it finds a model very similar in class abundance and parameter estimates to the true model. Therefore it appears that EM search is not spending enough time searching the six class model sub-space.

We tried our MCE sampler on the six Gaussian variate problem that EM search faired poorly on. With the same amount of computation time given to EM search, the best model our MCE sampler found had a message length of 57952 nits. This model has a shorter message length than the one found by EM search and the class abundances and parameter estimates are more similar to that of the generation mechanism's. We found this model with regularity from multiple independent annealing runs with a random number of initial classes, starting at temperature 2.0 and spending 50 iterations at each temperature. The cooling constant was 0.99. The model found is $e^{80}$ times more likely that the model found by EM search.

If we keep the same generation mechanism and sample sizes but increase the standard deviation of attributes to 0.6, i.e. $\sigma_{1..6,1..6}=0.6$ then EM search was able to find the correct number of classes but unable to find a good approximation to the generation mechanism's parameters, even when we told it the correct number of classes. The MCE sampler performs similarly as on the previous version of the problem. By increasing σ we have increased the amount of overlap between the classes.

We have also verified that in a trivial multi-modal problem that the sampler visits each mode with a chance (relative to the other modes) equal to its posterior probability. This did not occur when we tried randomly restarted EM search. Instead EM search would regularly converge to a subset of the modes. We feel this occurs because even though the modes have similar probabilities, their basins of attraction were different sizes.

## 10 CONCLUSION AND RELATED WORK

We have shown by making a minor and natural change to the EM and K-Means algorithms that a Gibbs sampler for a MML defined posterior can be constructed. Gibbs sampling unlike the previous two algorithms does not converge to a point estimator. We illustrated how this sampler can explore a posterior distribution of varying dimensionality. This removes limitations of the K-Means and EM algorithms we described such as estimating the number of classes to fit the data to.

Our purpose in this paper is to communicate how a small change can be made to the K-Means and EM algorithms that results in a method of overcoming their limitations. We have not discussed our current work in preparation that compares our approach to other more elaborate samplers that explore model spaces of varying dimensionality and using model space estimators such as BIC and AIC in combination with EM.

By sampling from a model space of varying dimension we obtain a more complete picture of the posterior distribution. We have used this benefit to handle problems in autonomous learning (Davidson 1998) by finding alternative explanations of the data. We demonstrated that our sampler finds better models than an EM algorithm and that in multi-modal problems visits each mode with a chance approximately equal to its posterior probability. We have perhaps portrayed MCMC too simply and note that MCMC samplers often suffer from problems of slow convergence to the posterior distribution and poor mixing (Gilks 1996). We have addressed these issues by making improvements (Davidson 1998) to the base sampler described in this paper.

## ACKNOWLEDGMENTS

I am most grateful to Emeritus Professor C.S. Wallace for his patience in explaining many aspects of MML induction and MCMC sampling. I would like to thank the anonymous reviewers for their valuable comments.